\documentclass[letterpaper, 10 pt, conference]{ieeeconf}
\IEEEoverridecommandlockouts                              

\usepackage{cite}
\usepackage{amsmath,amssymb,amsfonts}
\usepackage{algorithmic}
\usepackage{graphicx}
\usepackage{textcomp}
\usepackage{xcolor}
\usepackage{booktabs}

\makeatletter
\let\NAT@parse\undefined
\makeatother
\usepackage{hyperref}
\hypersetup{
    colorlinks=true,
    linkcolor=blue,
    filecolor=magenta,      
    urlcolor=cyan,
    citecolor=red,
    pdftitle={goslam},
    pdfpagemode=FullScreen,
}
\begin{document}

\title{HAVEN: Hierarchical Adversary-aware Visibility-Enabled Navigation with Cover Utilization using Deep Transformer Q-Networks\\
\thanks{}
}

\author{Mihir Chauhan$^{1}$, Damon Conover$^{2}$, Aniket Bera$^{1}$%
\thanks{\raggedright $^{1}$Mihir Chauhan, and Aniket Bera are with the IDEAS Lab, Department of Computer Science, Purdue University, West Lafayette, IN, USA. %
        {\tt\small \{\href{mailto:chauhanm@purdue.edu}{chauhanm}\allowbreak,\href{mailto:aniketbera@purdue.edu}{aniketbera}\allowbreak\}@\allowbreak purdue\allowbreak.edu}}
\thanks{\raggedright $^{2}$Damon Conover is with the DEVCOM Army Research Laboratory, Adelphi, MD, USA. %
        {\tt\small \{\href{mailto:chauhanm@purdue.edu}{damon.m.conover.civ@army.mil}\}}}%
}

\maketitle

\begin{abstract}
Autonomous navigation in partially observable environments requires agents to reason beyond immediate sensor input, exploit occlusion, and ensure safety while progressing toward a goal. These challenges arise in many robotics domains, from urban driving and warehouse automation to defense and surveillance. Classical path planning approaches and memory-less reinforcement learning often fail under limited fields-of-view (FoVs) and occlusions, committing to unsafe or inefficient maneuvers. We propose a hierarchical navigation framework that integrates a Deep Transformer Q-Network (DTQN) as a high-level subgoal selector with a modular low-level controller for waypoint execution. The DTQN consumes short histories of task-aware features, encoding odometry, goal direction, obstacle proximity, and visibility cues, and outputs Q-values to rank candidate subgoals. Visibility-aware candidate generation introduces masking and exposure penalties, rewarding the use of cover and anticipatory safety. A low-level potential field controller then tracks the selected subgoal, ensuring smooth short-horizon obstacle avoidance. We validate our approach in 2D simulation and extend it directly to a 3D Unity–ROS environment by projecting point-cloud perception into the same feature schema, enabling transfer without architectural changes. Results show consistent improvements over classical planners and RL baselines in success rate, safety margins, and time-to-goal, with ablations confirming the value of temporal memory and visibility-aware candidate design. These findings highlight a generalizable framework for safe navigation under uncertainty, with broad relevance across robotic platforms.
\end{abstract}

\section{Introduction}
Autonomous navigation in real-world environments requires agents to reason under uncertainty, avoid obstacles smoothly, and make steady progress toward a goal despite limited perception. These challenges appear across domains such as urban driving with occluded pedestrians, warehouse robots operating in cluttered aisles, and field robots moving through vegetation. In defense and surveillance applications, the problem is further complicated by \emph{adversarial agents} whose fields-of-view (FoVs) must be actively avoided. In such settings, exposure may lead to detection or failure, making navigation both a geometric and a tactical problem. Effective navigation therefore demands not only spatial reasoning but also the ability to exploit occlusion, anticipate visibility, and retain relevant past observations.

\begin{figure}[htbp]
    \centerline{\includegraphics[width=3.5in]{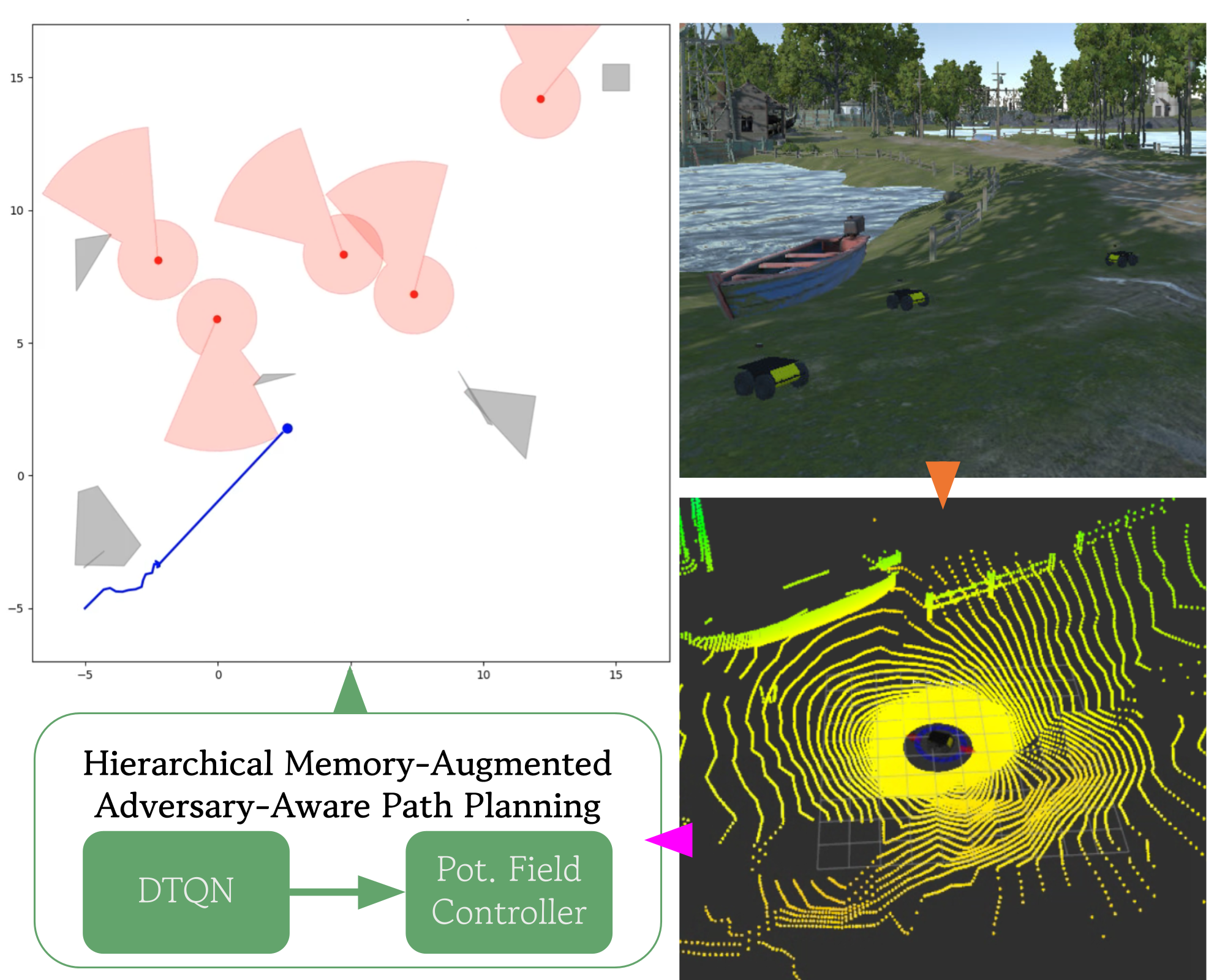}}
    \caption{Proposed hierarchical architecture for enemy-aware path planning with cover utilization using high-level Deep Transformer Q-Network and low-level potential field controller. }
    \label{fig:cover}
\end{figure}

Classical local planners such as the Dynamic Window Approach (DWA) and Vector Field Histogram (VFH) remain widely used for short-horizon obstacle avoidance. While effective in open environments, these approaches rely on instantaneous sensing and degrade significantly under limited FoVs and occlusions, often committing to unsafe or inefficient maneuvers (see Sec.~VII). Reinforcement learning (RL) methods have shown promise for navigation, but memory-less variants struggle in partially observable domains: when crucial state information lies outside the current sensor view, such policies fail to leverage cover, anticipate visibility, or reason about hidden threats. These limitations motivate frameworks that combine geometric safety priors with memory-enabled decision making.

Recent work in RL under partial observability has introduced recurrent and Transformer-based architectures, which capture temporal dependencies over long horizons. However, their application to robotic navigation remains limited, particularly for visibility-constrained and adversarial scenarios. In parallel, hierarchical control has proven effective in decomposing complex navigation tasks into high-level decision making and low-level control. Yet most existing approaches focus on geometric safety and efficiency alone, without explicitly modeling enemy visibility or adversarial occlusion-aware reasoning.

In this work, we address these gaps with a hierarchical, memory-augmented navigation framework that integrates a Deep Transformer Q-Network (DTQN) as a high-level subgoal selector with a low-level controller for waypoint execution. Our contributions are threefold:
\begin{enumerate}
    \item \textbf{Visibility- and adversarial-aware subgoal selection:} We design compact, task-aware feature vectors that encode geometry, goal progress, obstacle distances, and adversarial visibility cues, enabling the high-level DTQN to learn occlusion- and cover-aware strategies.
    \item \textbf{Hierarchical integration with low-level control:} The DTQN selects subgoals from a candidate set, while a potential field controller executes short-horizon trajectories, ensuring smooth obstacle avoidance, reactivity, and evasive maneuvers against adversarial agents.
    \item \textbf{Unified 2D--3D extension:} We demonstrate transfer by extending the 2D framework to 3D Unity--ROS environments, projecting point-cloud perception into the same feature schema, enabling direct reuse of the trained DTQN without architectural changes.
\end{enumerate}

We evaluate our framework in both 2D and 3D environments, showing consistent improvements over classical planners and RL baselines in terms of success rate, obstacle clearance, reduced exposure to adversarial FoVs, and time-to-goal. Ablation studies confirm the importance of temporal memory, candidate generation, and visibility-aware design.

The remainder of this paper is organized as follows. Section~II reviews related work on memory-based RL, local planning under partial observability, and hierarchical navigation. Section~III formulates the problem as a Partially Observable Markov Decision Process (POMDP) with adversarial visibility constraints. Section~IV presents the hierarchical DTQN framework. Section~V describes the experimental setup, and Section~VI reports results and ablations. Sections~VII and VIII provide analysis and discussion. Section~IX outlines future research directions, and Section~X concludes the paper.

\section{Related Work}

\subsection{Approaches for Collision Avoidance}

Collision avoidance has been a long-standing problem in robotics and autonomous navigation, with a variety of approaches proposed in both classical and reinforcement-learning methods. Classical methods often rely on geometric reasoning and reactive control laws, such as velocity obstacles introduced by Fiorini and Shiller \cite{doi:10.1177/027836499801700706} which computes admissible accelerations that prevent collisions with moving obstacles. More recent approaches utilize reinforcement learning, such as CoverNav, in which Hossain et al. \cite{hossain2023covernav} navigates outdoor environments by following cover, determined by visual perception models and LIDAR point clouds. Wang et al. \cite{9461231} deploy a multi-level autonomous vehicle control schema, using reinforcement learning for submodules that are merged into a final steering and brake control policy. Similarly, Lin et al. \cite{lin2023occlusion} solve path planning with partial observability, while Guvenkaya et al. \cite{guvenkaya2024local} use predetermined waypoints to plan paths with collision aware sensing with an extrapolation model.

\subsection{Adversarial Threat-Aware Navigation}

Adversarial Threat-Aware Navigation is relatively less explored compared to collision avoidance in path planning, but approaches, such as Hossain et al. \cite{hossain2024encomp} utilize offline reinforcement learning to enable robots to navigate in cover while being penalized for being exposed to potential targets, using a Conversative Q-Learning model, needing a relatively large dataset for training. Auode et al. \cite{5650734} approach similar problems except with the goal of stray object avoidance where threats are assessed based on their real-time motion into an autonomous vehicle’s path, where intentions of such moving objects are predicted and then assessed as threats as needed.

\subsection{Hierarchical Reinforcement Learning Control Strategies}

Hierarchical reinforcement learning (HRL) has been widely explored as a means to decompose complex decision-making into structured levels of abstraction, specifically applications of autonomous path planning in dynamic traffic scenarios. Zhang et al. \cite{zhang2025trajectory} utilize a upper level deep reinforcement model with a lower level controller using model predictive control (MPC). The upper level model plans high level policies which are implemented with an MPC controller for tasks of driving unmanned vehicles in traffic. Similar approaches, such as Naveed et al. \cite{9564634} utilize a 3 layer approach with high-level policy options and a Long-Short-Term-Memory layer to help with observation noise. Other approaches such as Zheng et al. \cite{7378017} propose using HRL for multi-agent planning to reduce the state space and divide learning into higher and lower level tasks, while Lu et al. \cite{9241055} incorporate a kernel-based least-squares policy iteration with uneven sampling and pooling to efficiently solve decision-making problems on a high level.

\subsection{Deep Transformer Q-Network}

In this paper, we discuss the usage of Deep Transformer Q-Networks (DTQN) for the purpose of path planning with collision and adversarial threat avoidance, which to our knowledge is a novel application of this architecture to a less explored problem. Esslinger et al. \cite{esslinger2022deep} have shown that DTQNs are effective in tasks involving partial observability, where memory of prior observations helps improve performance over recurrent neural networks (RNNs). Hence, we chose to use this model architecture as a high-level subgoal selector model. For the development of our model, we modified the source code provided by Esslinger et al. https://github.com/kevslinger/DTQN.

\section{Problem}
We formulate the task of adversarially aware navigation with cover utilization as a Partially Observable Markov Decision Process (POMDP), following \cite{esslinger2022deep,KAELBLING199899}. In this setting, the agent is not given the full state of the environment at each timestep but must reason under limited observability. The goal is to reach a target location while avoiding collisions and minimizing exposure to adversarial agents’ fields-of-view (FoVs). This formulation captures both the geometric challenges of obstacle avoidance and the tactical requirement of occlusion-aware motion.

Formally, the POMDP is defined as the tuple
\[
\mathcal{M} = (\mathcal{S}, \mathcal{A}, \mathcal{O}, T, Z, R, \gamma),
\]
where $\mathcal{S}$ is the state space, $\mathcal{A}$ the action space, $\mathcal{O}$ the observation space, $T$ the state transition function, $Z$ the observation model, $R$ the reward function, and $\gamma \in [0,1)$ the discount factor.

\subsection{State Space}
At time $t$, the environment state $s_t \in \mathcal{S}$ includes:
\begin{itemize}
    \item Agent pose and velocity: $(x_t, y_t, v_{x,t}, v_{y,t})$
    \item Static obstacle and occluder geometry in 2D (polygonal map)
    \item Enemy positions, headings, and FoV parameters
    \item Goal position $g^\star$
\end{itemize}

\subsection{Observation Model}
The agent cannot observe $s_t$ directly. Instead, at each timestep it receives a partial observation $o_t \in \mathcal{O}$ given by
\begin{equation}
    o_t \sim Z(o_t \mid s_t, a_{t-1}),
\end{equation}
where $a_{t-1} \in \mathcal{A}$ is the previous action, and $Z$ encodes the FoV constraint and line-of-sight occlusions. Thus, $Z$ maps the true environment into the subset of features perceivable by the agent’s sensors.

Each $o_t$ is represented as a structured feature vector including:
\begin{itemize}
    \item \textbf{Relative obstacle and subgoal positions:} coordinates of visible obstacles and candidate subgoals in the agent’s local frame.  
    \item \textbf{Enemy visibility features:} indicators for whether an enemy has line-of-sight to the agent, and the minimum distance to visible enemy FoVs.  
    \item \textbf{Relative goal information:} vector displacement to the goal if within the FoV; otherwise, this term is masked.  
\end{itemize}

The high-level policy consumes an observation history 
$h_t = (o_{t-k+1}, \ldots, o_t)$ 
to incorporate memory across the past $k$ steps, enabling temporal reasoning about occlusion and threat exposure.

\subsection{Reward Function}
The reward function penalizes unproductive or unsafe behavior—such as lingering near enemy FoVs, revisiting prior subgoals, or colliding with obstacles—while rewarding productive subgoal progress and shorter time-to-goal. The exact structure of the reward is described in Section~V.

\subsection{Candidate Subgoal Features}
Each candidate subgoal $g_t^i \in \mathcal{G}_t$ is represented by a feature vector $f_t^i \in \mathbb{R}^{16}$:
\begin{align*}
  f_t^i = \big[ & x_t, y_t, v_{x,t}, v_{y,t}, (g^\star_x - x_t), (g^\star_y - y_t), d_{\text{goal}}, \\
  & g^i_x, g^i_y, (g^i_x - x_t), (g^i_y - y_t), d_{\text{cand}}, n_{\text{enemy}}, d_{\min}^{\text{enemy}}, \\
  & n_{\text{vis}}, \mathbb{1}_{\text{seen}} \big].
\end{align*}
Here $(x_t, y_t)$ and $(v_{x,t}, v_{y,t})$ denote the agent state; $(g^\star_x - x_t, g^\star_y - y_t)$ the goal displacement with $d_{\text{goal}}$ its norm; $(g^i_x, g^i_y)$ the candidate centroid; $(g^i_x - x_t, g^i_y - y_t)$ the agent-to-candidate vector with $d_{\text{cand}}$ its norm; $n_{\text{enemy}}$ the number of adversaries; $d_{\min}^{\text{enemy}}$ the minimum candidate-to-enemy distance; $n_{\text{vis}}$ the number of enemies with line-of-sight to the candidate; and $\mathbb{1}_{\text{seen}}$ indicates whether the agent is currently exposed.

\subsection{Evaluation Metrics}
We assess performance with:
\begin{itemize}
    \item \textbf{Success Rate:} fraction of runs reaching $g^\star$  
    \item \textbf{Time-to-Goal:} steps to reach $g^\star$  
    \item \textbf{Path Length:} total trajectory distance  
    \item \textbf{Safety Margin:} minimum clearance to obstacles  
    \item \textbf{Visibility Exposure:} cumulative time exposed to enemy FoVs  
    \item \textbf{Collision Rate:} fraction of runs terminated by collision  
\end{itemize}
Together, these metrics evaluate efficiency, robustness, and stealth—critical for navigation in adversarial and visibility-constrained environments.

\section{Methodology}
\begin{figure}[htbp]
    \centerline{\includegraphics[width=3.5in]{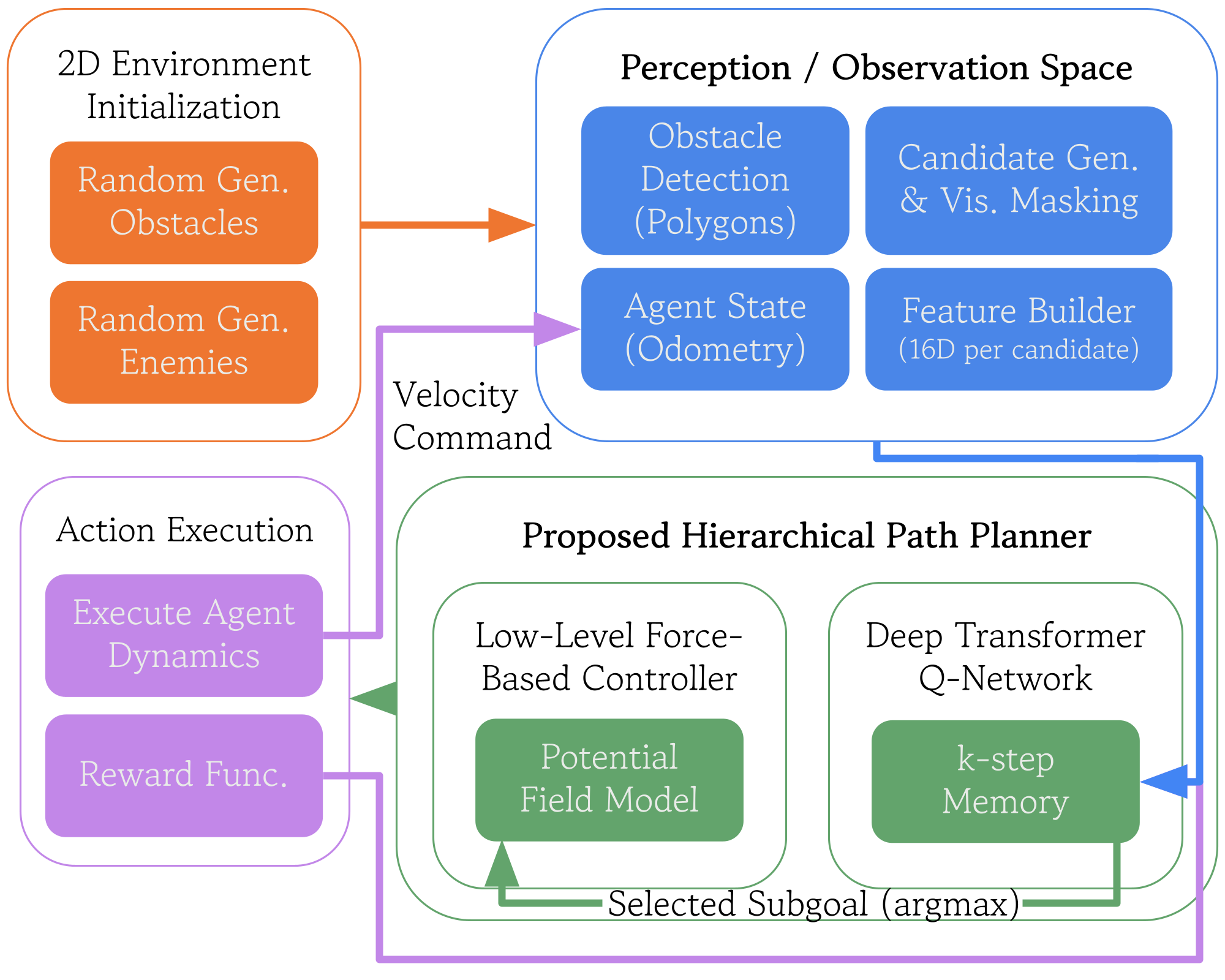}}
    \caption{Proposed hierarchical architecture. The high-level DTQN outputs Q-values for candidate subgoals, and the selected subgoal is passed to the low-level potential field controller for execution.}
    \label{fig:framework}
\end{figure}

We propose a hierarchical decision-making framework that enables an agent to reach a goal while avoiding static obstacles and minimizing exposure to adversarial fields-of-view (FoVs). The approach integrates a high-level policy for subgoal selection with a low-level controller for reactive execution. An overview is shown in Fig.~\ref{fig:framework}.

\subsection{Hierarchical Policy Structure}
At each timestep $t$, the agent executes two levels of control:
\begin{enumerate}
    \item \textbf{High-Level Policy $\pi_H$:} selects a subgoal $g_t$ from a candidate set $\mathcal{G}_t$ using the history of partial observations $h_t$.
    \item \textbf{Low-Level Controller $\pi_L$:} generates continuous control actions $a_t \in \mathcal{A}_L$ that drive the agent toward $g_t$ while performing local obstacle avoidance and evasive maneuvers.
\end{enumerate}
This decomposition enables long-horizon planning at the subgoal level while ensuring short-horizon safety and reactivity.

\textcolor{black}{\textbf{Advantages over end-to-end methods.} Our hierarchical design offers three key benefits. First, it reduces the effective action space: the high-level policy selects from $|\mathcal{G}_t| \leq 9$ discrete goals instead of outputting continuous 2D velocities at every timestep, improving sample efficiency. Second, safety-critical behaviors—obstacle avoidance, boundary enforcement, and enemy evasion—are handled by a deterministic low-level controller with repulsive potential fields, avoiding the need to learn reactive collision avoidance from sparse rewards. Third, the modular separation enables direct transfer to 3D (Sec.~VI-B): 2D force vectors are mapped to $(v, \omega)$ ROS commands, and the high-level DTQN weights are reused without retraining in most cases.}

\subsection{Observation Model}
Instead of raw sensor inputs, each candidate subgoal $g_t^i \in \mathcal{G}_t$ is represented by a compact feature vector $f_t^i \in \mathbb{R}^{16}$:
\begin{itemize}
    \item \textbf{Agent state:} $(x_t, y_t, v_{x,t}, v_{y,t})$
    \item \textbf{Goal-relative terms:} $(g^\star_x - x_t,\, g^\star_y - y_t)$ and $d_{\text{goal}}$
    \item \textbf{Candidate geometry:} $(g^i_x, g^i_y)$, $(g^i_x - x_t,\, g^i_y - y_t)$ and $d_{\text{cand}}$
    \item \textbf{Enemy visibility states:} number of enemies $n_{\text{enemy}}$, minimum enemy-to-candidate distance $d_{\min}^{\text{enemy}}$, number of enemies with line-of-sight $n_{\text{vis}}$, and an indicator $\mathbb{1}_{\text{seen}}$ if the agent is currently exposed
\end{itemize}
Obstacle and occluder geometry are obtained from the 2D polygonal map (or from point-clouds in 3D). Enemy line-of-sight is computed against these polygons. The DTQN input for candidate $i$ is a length-$k$ sequence $X_t^i \in \mathbb{R}^{k \times 16}$ formed by tiling $f_t^i$, enabling temporal reasoning.

\subsection{Candidate Subgoal Generation}
At each timestep, candidate subgoals are sampled within a local horizon from regions of free space:
\begin{equation}
g_t^i \in \mathcal{F}_t = \{p \in \mathbb{R}^2 \mid p \notin \mathcal{O},~p \notin \text{FoV}(E)\},
\end{equation}
where $\mathcal{O}$ are static obstacles and $\text{FoV}(E)$ is the union of enemy FoV sectors. This biases candidates toward occluded and safer regions.

\subsection{Feature Encoding for Subgoal Evaluation}
Each candidate $g_t^i$ is encoded into a tactical feature vector:
\begin{equation}
f(g_t^i) = [d_{\text{goal}}, d_{\text{obs}}, d_{\text{FoV}}, \theta_{\text{align}}, \ldots],
\end{equation}
where $d_{\text{goal}}$ is the distance to the goal, $d_{\text{obs}}$ the obstacle clearance, $d_{\text{FoV}}$ the distance to enemy visibility, and $\theta_{\text{align}}$ the heading alignment. All features are normalized and concatenated into a fixed-length representation.

\subsection{High-Level Subgoal Selection}
For each candidate $g_t^i$, the DTQN processes the sequence $X_t^i$ to produce Q-values:
\begin{equation}
Q_\theta(g_t^i \mid h_t) := \big[\mathrm{DTQN}_\theta(X_t^i)\big]_{k,1}.
\end{equation}
The chosen subgoal is selected $\varepsilon$-greedily during training:
\begin{equation}
g_t =
\begin{cases}
\text{Uniform}(\mathcal{M}_t), & \varepsilon,\\[4pt]
\arg\max_{i \in \mathcal{M}_t} Q_\theta(g_t^i \mid h_t), & 1-\varepsilon,
\end{cases}
\end{equation}
where $\mathcal{M}_t$ is the valid candidate set after masking. At test time, $\varepsilon \!\to\! 0$ and selection is greedy.

\textcolor{black}{To reduce oscillatory subgoal switching---where the policy alternates between candidates at consecutive decisions---we introduce a hysteresis bonus $\beta_H = 0.15$ that is added to the Q-value of the previously selected subgoal before the argmax. This encourages the agent to commit to a subgoal for multiple decision steps unless a substantially better alternative emerges, producing smoother trajectories and more decisive cover utilization.}

\subsection{Reward Function}
The reward balances progress, safety, and stealth:
\begin{equation}
r_t = w_p (d_{t-1}^{\text{sub}} - d_{t}^{\text{sub}})
      - w_e \,\mathbb{1}_{\text{exposed}_t}
      - w_c \,\mathbb{1}_{\text{collision}_t}
      - w_t.
\end{equation}
Bonuses are added for reaching a subgoal or the final goal:
\begin{equation}
r_t \leftarrow r_t +
\begin{cases}
b_{\text{subgoal}}, & \text{if subgoal reached},\\
b_{\text{goal}}, & \text{if goal reached},\\
\textcolor{black}{p_{\text{fail}},} & \textcolor{black}{\text{if episode timeout},}\\
0, & \text{otherwise}.
\end{cases}
\end{equation}

\subsection{Low-Level Execution}
The low-level controller executes the selected subgoal using potential field forces that combine attraction, obstacle repulsion, adversary avoidance, and escape behaviors:
\begin{equation}
F_{\text{tot}}(\tau) = w_d F_{\text{des}} + w_o F_{\text{obs}} + w_e F_{\text{enemy}}
+ w_a F_{\text{ant}} + w_s F_{\text{esc}}.
\end{equation}
The resulting velocity command is smoothed and clipped by $v_{\max}$:
\begin{equation}
u_\tau = \beta v_\tau + (1-\beta)F_{\text{tot}}(\tau), \quad
v_{\tau+1} = \mathrm{sat}_{v_{\max}}(u_\tau).
\end{equation}
This hybrid design ensures tactical subgoal reasoning at the high level while maintaining reactive, safe execution at the control level.

\subsection{Training Procedure}
We train the high-level policy $\pi_H$ with reinforcement learning using the reward above. The low-level controller $\pi_L$ is rule-based and parameter-tuned. This division keeps the state space manageable while enabling the high-level DTQN to focus on temporal and adversarial reasoning, and the low-level controller to guarantee safety.

\textcolor{black}{To ensure stable Q-learning, the replay buffer stores full transitions $(h_t, R_t, \gamma^k, h_{t+1}, \text{done})$ rather than pre-computed TD targets to avoid stale-targets where pre-computed values become inconsistent as the policy evolves. Also, we use \emph{Double DQN} \cite{10.5555/3016100.3016191} where the online network selects the best next action $a^* = \arg\max_a Q_\theta(h_{t+1}, a)$, while the target network evaluates it as $Q_{\theta^-}(h_{t+1}, a^*)$, reducing Q-value overestimation that could cause a policy to confidently select poor subgoals.}

\section{Experimental Setup}
We evaluate our hierarchical framework in both 2D and 3D simulated environments that capture the challenges of stealth navigation in adversarial settings. The experiments measure how effectively the agent balances goal-directed progress, obstacle avoidance, and minimization of exposure to enemy FoVs.

\begin{figure}[htbp]
    \centerline{\includegraphics[width=3.5in]{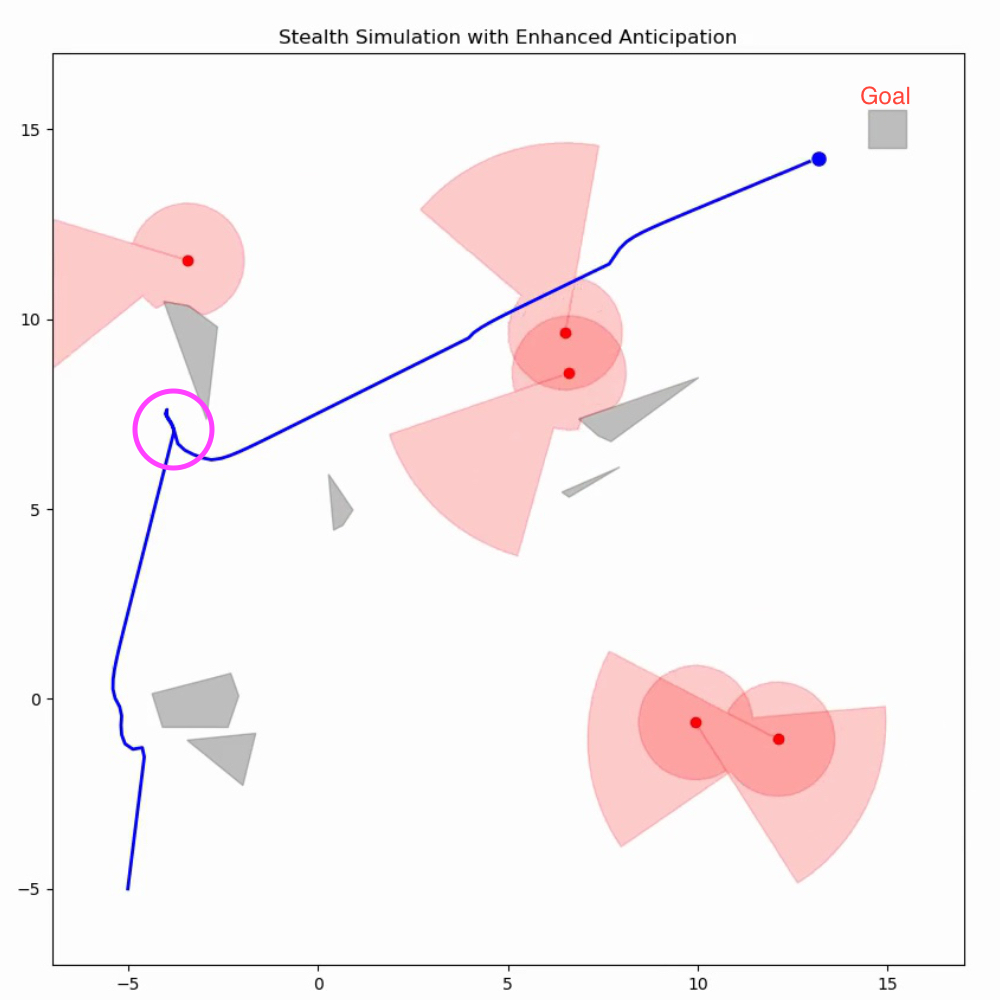}}
    \caption{2D training and evaluation setup with 5 enemy agents (red) and the navigating agent (blue). The agent’s trajectory avoids exposure by using cover (obstacle circled in pink) before reaching the goal (grey square at (15,15)). Enemy spawn points and obstacles are randomized at the start of each run.}
    \label{fig:2d-environment}
\end{figure}

\subsection{2D Environment Design}
We consider a continuous square workspace of size $20 \times 20$\,m, with bounds $[x_{\min},x_{\max}] \times [y_{\min},y_{\max}]$ where $(x_{\min},y_{\min})=(-5,-5)$ and $(x_{\max},y_{\max})=(15,15)$.  

\textbf{Obstacles.} Static obstacles, which also serve as occluders, are generated as random convex polygons:
\begin{itemize}
    \item \emph{Polygon generation:} each obstacle is defined by a random center $(c_x,c_y)$, radius $r \!\sim\! \mathcal{U}(0.5,2.0)$\,m, and $v\!\in\!\{3,\ldots,5\}$ vertices projected at random angles.
    \item \emph{Quantity and clutter:} the number of obstacles $N_{\text{obs}}$ is varied to create open versus cluttered maps, with larger $N_{\text{obs}}$ or radii yielding narrower passages.
\end{itemize}

\textbf{Enemies.} Adversarial agents are spawned at random interior positions:
\begin{itemize}
    \item \emph{Kinematics:} each enemy moves at constant speed with small random heading changes, bouncing off boundaries and avoiding obstacles with short-range repulsion.
    \item \emph{Sensing model:} each enemy has a sector-shaped FoV with aperture $\theta_{\text{FoV}}$ and range $r_{\text{FoV}}$, augmented by a reduced-range $360^\circ$ ring. Ray-casting against obstacles yields non-convex visibility regions (see Fig.~\ref{fig:2d-environment}).
\end{itemize}

The agent begins near the lower-left corner and must reach a goal near $(x_{\max}-1, y_{\max}-1)$. Training maps are diversified by randomizing $N_{\text{obs}}$, obstacle radii, enemy count, and headings. Occlusions from polygonal obstacles naturally create cover opportunities.

\subsection{Training Details}
The high-level policy is trained as a DTQN using temporal-difference learning with mean-squared error loss. After up to $k$ low-level steps per high-level decision, the return target is:
\[
R_t = \sum_{j=0}^{k-1} \gamma^j r_{t+j},
\]
\textcolor{black}{
\begin{equation}
\hat{Q}_t =
\begin{cases} 
R_t, & \text{terminal} \\
R_t + \gamma^k Q_{\theta^-} \Big( h_{t+1}, & \\ 
\quad \arg\max_{i} Q_\theta(g_{t+1}^i \mid h_{t+1}) \Big), & \text{otherwise}
\end{cases}
\end{equation}
where the Double DQN formulation uses the online network $Q_\theta$ to select the best next candidate and the target network $Q_{\theta^-}$ to evaluate it, reducing overestimation bias.
}
We minimize $\tfrac{1}{2}\big(Q_\theta(g_t^{i^*} \mid h_t) - \hat{Q}_t\big)^2$ with Adam (lr $=10^{-3}$) and $\varepsilon$-greedy exploration \textcolor{black}{($\varepsilon$ decaying from 1.0 to 0.08)}. We use $k=3$ (causal attention with positional encodings) and re-evaluate after at most \textcolor{black}{$H=20$} low-level steps. Candidate subgoals are sampled from obstacle centroids and the goal, with heuristics to reject near-duplicate or excessively distant points. Training proceeds online with periodic checkpointing.

\subsection{Comparison Methods}

We compare against the following baselines, implemented within the same simulation stack, geometry, sensing/FoV models, and episode protocol (identical environment generator, seeds, and metrics) to ensure comparable conditions:
\begin{itemize}
    \item \textbf{Potential Fields + Predicted FoV Unions:} A purely reactive controller combining goal attraction, obstacle repulsion, and an anticipatory repulsion from the union of predicted enemy FoVs, which is being used as the low level controller in HAVEN. The force field is computed from the same polygonal obstacles and enemy FoV geometry used for all methods, with identical smoothing and speed constraints in the dynamics integrator.
    \item \textbf{LSTM (end-to-end):} An LSTM policy that outputs a direct 2D control vector at each high-level step based on compact task-aware features (agent pose/velocity, goal delta/distance, nearest obstacle clearance, enemy visibility, and enemy heading). Actions are integrated by the same dynamics model and subject to the same speed caps and smoothing; no MPC is used.
    \item \textbf{LSTM HL + MPC:} A hierarchical variant where an LSTM serves as the high-level subgoal selector over the same candidate set as HAVEN (obstacle centroids plus goal with identical masking). The chosen subgoal is tracked by the same MPPI-style MPC and low-level dynamics as in our framework, using identical cost weights and limits. The LSTM Model used in this and the previous were the same, independently trained model, as we refrained from using the DTQN as a teacher since independent comparisons were desired. Training was similar in episode count to that of the DTQN used in HAVEN.  
    \item \textbf{VFH+ (FoV-limited):} A histogram-based local planner constructed from the same obstacle polygons, augmented with penalties for heading bins that would place the agent within current enemy visibility polygons. Control outputs are integrated with the same smoothing and speed limits.
    \item \textbf{Visibility-Aware Greedy:} Heading sampling with a cost that combines distance-to-goal, obstacle clearance penalties, and a binary exposure term based on the current enemy FoV union. Evaluated under the same time step and integration scheme.
    \item \textbf{DWA (FoV-limited):} Dynamic Window Approach sampling linear/angular velocities and rolling out short-horizon dynamics with costs for goal progress, obstacle proximity, and enemy FoV exposure. Uses the same state integrator and limits.
    \item \textbf{HAVEN, Memory-Less:} Our hierarchy with DTQN memory ablated by setting $k{=}1$ while keeping the same feature schema, candidate set, masking, and low-level controller.
    \item \textbf{HAVEN (proposed):} Our full hierarchical DTQN high-level selector with the low-level controller.
\end{itemize}
\subsection{Evaluation Protocol}
We generate $100$ randomized test environments with varied obstacle layouts and enemy placements. Each method is run for $50$ episodes per environment, and results are averaged. Metrics include success rate, path length, time-to-goal, collision rate, and cumulative exposure time, capturing efficiency, robustness, and stealth in adversarial navigation.

\section{Results}

\subsection{Quantitative Comparisons to Baselines}
We evaluate the proposed hierarchy against five baselines/ablations over $100$ randomized 2D environments (common seed). Metrics include \emph{success rate}, \emph{collision rate}, \emph{path length}, \emph{time-to-goal}, and \emph{exposure time} (average steps under enemy FoV). Results are summarized in Table~\ref{tab:quantitative_results}.

The hierarchical method achieves a near-perfect success rate (99\%) and the lowest exposure (2.38), outperforming all baselines. Removing temporal memory degrades safety and stealth (collision $0.20$ vs.\ $0.04$, exposure $4.20$ vs.\ $2.38$), indicating that short history improves cover utilization and reduces risky commitments. Classical planners (VFH, DWA) attain shorter paths and/or times but at substantially higher collision rates (52\% and 40\%, respectively) and elevated exposure, reflecting an unfavorable safety--speed trade-off in visibility-constrained settings. The visibility-aware greedy policy keeps exposure relatively low but is myopic, lowering success (90\%) and increasing collisions (0.22). A purely reactive potential-field variant with predicted FoV unions minimizes collisions (0.02) but suffers poor success (0.49) and higher exposure, highlighting the need for high-level temporal reasoning.

\begin{table*}[t]
    \centering
    \caption{Quantitative results across $100$ randomized environments. Best (or tied best) values per column are in \textbf{bold}.}
    \label{tab:quantitative_results}
    \resizebox{0.9\textwidth}{!}{
    \begin{tabular}{lccccc}
        \toprule
        \textbf{Method} & \textbf{Success} & \textbf{Collision} &
        \textbf{Path Length} & \textbf{Time} & \textbf{Enemy Exposure} \\
        \midrule
        HAVEN (proposed) & \textbf{0.97} & \textbf{0.04} & 70.7 & 31.0 & \textbf{2.38} \\
        HAVEN, Memory-Less & 0.90 & 0.20 & 82.7 & 39.8 & 4.20 \\
        Potential Fields + Predicted FoV Unions & 0.49 & \textbf{0.02} & 37.5 & 34.1 & 5.98 \\
        EnCOMP & 0.42 & 0.04 & 88.4 & 26.9 & 3.67 \\
        LSTM (end-to-end) & 0.05 & 0.15 & 105.8 & 18.8 & 3.41 \\
        LSTM HL + MPC & 0.30 & 0.70 & 56.2 & 16.6 & 3.49 \\
        VFH & 0.97 & 0.52 & 32.2 & 25.1 & 2.81 \\
        Visibility-Aware Greedy & 0.90 & 0.22 & 40.4 & 38.3 & 2.60 \\
        DWA & 0.84 & 0.40 & \textbf{28.3} & \textbf{15.3} & 17.23 \\
        \bottomrule
    \end{tabular}
    }
\end{table*}

\begin{figure}[htbp]
    \centerline{\includegraphics[width=3.5in]{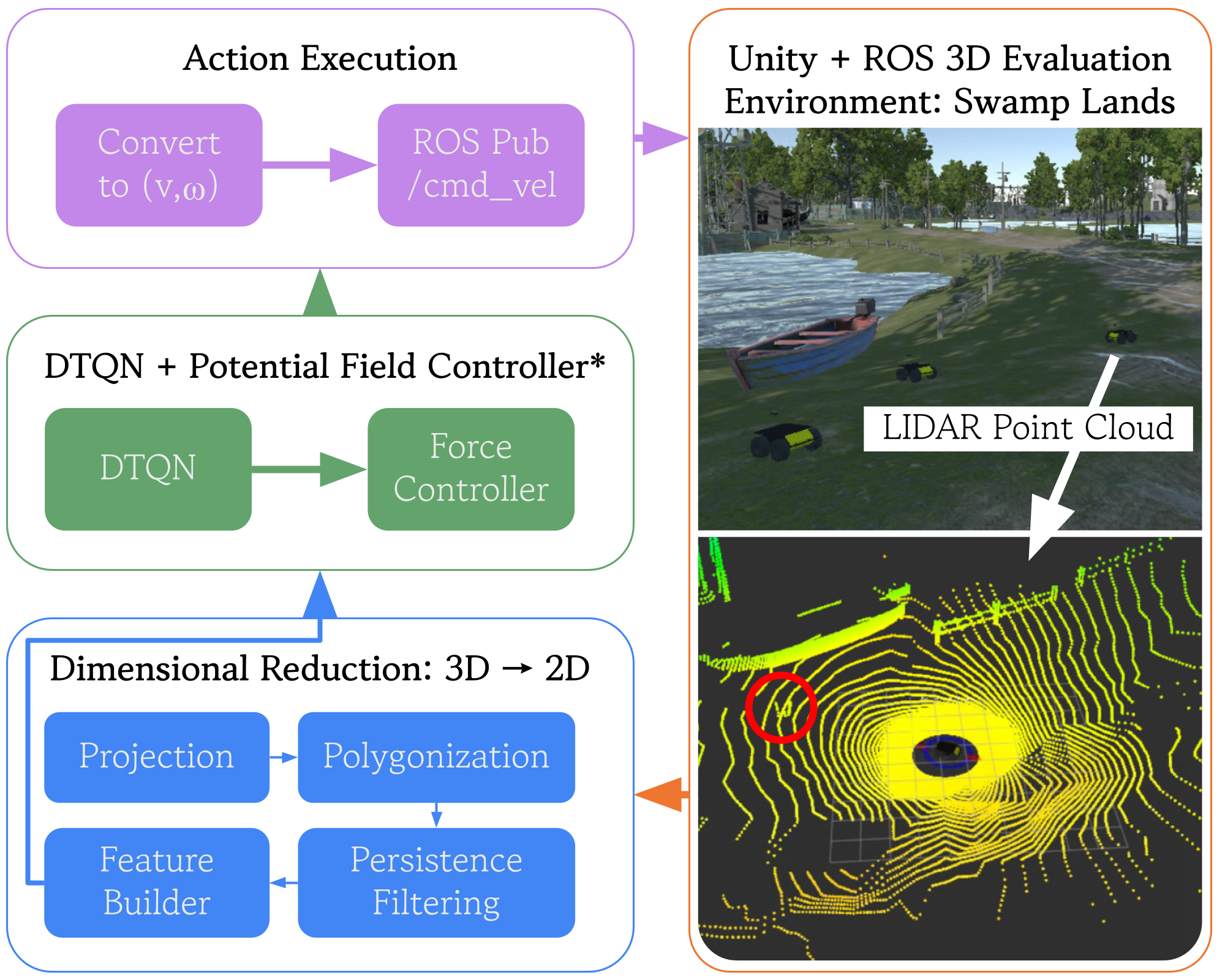}}
    \caption{3D evaluation pipeline. A 2D-trained DTQN + low-level controller are deployed in Unity--ROS: LiDAR point clouds are projected to the ground plane, denoised via occupancy/persistence, polygonized, and fed to the 2D feature schema. Selected subgoals are executed as linear/angular velocity commands in ROS. The red circle depicts an adversarial threat in the point cloud.}
    \label{fig:3d-extensions}
\end{figure}

\subsection{Extensions to 3D}
\textbf{Unity--ROS integration and point-cloud input.}
We deploy the 2D-trained DTQN unchanged in a 3D Unity--ROS environment by projecting perception to the ground plane and preserving the 2D feature schema (Fig.~\ref{fig:3d-extensions}). At each control step, the robot receives a point cloud $\mathcal{P}=\{p_i \in \mathbb{R}^3\}$ and odometry $(x,y,\theta)$ from ROS, forms a 2D obstacle set by planar projection and polygonization,
\begin{align}
\mathcal{S} &= \{(p_i^x,p_i^y):\, p_i\in\mathcal{P}\}, \\
\{\mathcal{O}_j\} &= \mathrm{PolyGen}(\mathcal{S}),
\end{align}
and applies persistence filtering: an obstacle $\mathcal{O}_j$ is kept only if observed in at least $T_{\mathrm{pers}}$ recent frames.

\paragraph{Feature and candidate construction (2D schema).}
From $\{\mathcal{O}_j\}$ we build $\mathcal{G}_t=\{g_t^i\}$ using obstacle centroids and the goal, masking near-duplicates and candidates beyond the agent-to-goal distance. For each valid $g_t^i$, we compute the same 16-D feature vector $f_t^i$ (agent pose/velocity, goal delta/distance, candidate geometry/distance, and adversarial visibility statistics), tile to $X_t^i\in\mathbb{R}^{k\times 16}$, and score with the last DTQN output.

\paragraph{High-level scoring and selection.}
Given $Y_t^i=\mathrm{DTQN}_\theta(X_t^i)\in\mathbb{R}^{k\times 1}$, we set $Q_\theta(g_t^i\!\mid h_t)=[Y_t^i]_{k,1}$ and select
\begin{equation}
g_t =
\begin{cases}
\mathrm{Uniform}(\mathcal{M}_t), & \text{w.p. }\varepsilon,\\
\arg\max_{i\in\mathcal{M}_t} Q_\theta(g_t^i\!\mid h_t), & \text{w.p. }1-\varepsilon,
\end{cases}
\end{equation}
with $\mathcal{M}_t$ the post-mask index set. When enabled, online TD updates match the 2D targets.

\paragraph{Enemies and visibility in 3D.}
Enemy agents are simulated in Unity and read via ROS topics. We compute line-of-sight on the 2D map using a sector FoV (aperture $\theta_{\mathrm{FoV}}$, range $r_{\mathrm{FoV}}$) plus a reduced-range $360^\circ$ ring. Ray-casting against $\{\mathcal{O}_j\}$ yields non-convex visibility polygons used both for features ($n_{\mathrm{vis}}$, $\mathbb{1}_{\mathrm{seen}}$, $d_{\min}^{\mathrm{enemy}}$) and for anticipatory safety (predicted FoV unions as temporary obstacles).

\paragraph{Low-level command synthesis (ROS).}
Let $p=(x,y)$ be the agent pose in the map frame and $\theta$ its heading. We compose $F_{\mathrm{tot}}$ (goal attraction, obstacle and FoV avoidance, escape) and convert to linear/angular rates:
\begin{align}
v &= \|F_{\mathrm{tot}}\|, \\
\theta_F &= \operatorname{atan2}(F_{\mathrm{tot}}^y, F_{\mathrm{tot}}^x), \quad
\Delta\theta = \mathrm{wrap}(\theta_F-\theta), \\
\omega &= \mathrm{sat}_{\omega_{\max}}\!\left(\frac{\Delta\theta}{\Delta t}\right).
\end{align}
Commands $(v,\omega)$ are published to ROS; the next state $(x’,y’,\theta’)$ is read from odometry. To accommodate computation/actuation latency, actions are held for several control periods before the next point-cloud update.

\paragraph{Practical considerations.}
We use ROS transforms to maintain a consistent 2D map frame. Persistence thresholds improve mapping stability under sparse/partial scans. Crucially, the hierarchical stack, feature schema, and DTQN weights are reused from 2D \emph{without architectural changes}, demonstrating direct transfer to 3D perception and control.

\textcolor{black}{The 2D-to-3D transfer relies on two main assumptions.
(1) \textbf{Planar environment:} Projecting LiDAR points $(p_i^x, p_i^y, p_i^z) \mapsto (p_i^x, p_i^y)$ discards elevation, which is valid for ground-level navigation with vertical obstacles but may fail in multi-layered settings or when adversarial FoVs pass above obstacles.
(2) \textbf{Enemy dynamics:} Adversarial agents in 3D are read from ROS topics using the same kinematics and FoV model as in 2D, assuming FoVs remain approximated by planar sectors.
Despite these constraints, the results suggest that temporal reasoning over visibility-aware subgoals is largely decoupled from perception, and the modular design allows targeted front-end upgrades (e.g., replacing $\mathrm{PolyGen}$ with a learned 3D-to-2D encoder) without retraining the high-level policy.}

\section{Analysis}
Table~\ref{tab:quantitative_results} shows that the proposed hierarchical DTQN consistently outperforms baselines in success, safety, and adversarial exposure. The memory-less variant exhibits higher collision (0.20) and exposure (4.20), confirming that temporal reasoning is crucial under partial observability.  

Classical planners show an expected advantage in efficiency (shorter paths and times) but degrade under adversarial constraints. DWA achieves the shortest time-to-goal (15.3) and path length (28.3), but with high collision (0.40) and exposure (17.23), reflecting an unfavorable safety–speed trade-off. VFH is similarly competitive in efficiency but unstable, with a 0.52 collision rate.  

The visibility-aware greedy baseline reduces exposure (2.60) but remains volatile (success 0.90, collision 0.22) due to its myopic, one-step decisions. Reactive potential fields with predicted FoV unions yield a low collision rate (0.02) but poor success (0.49) and moderate exposure (5.98), highlighting the limitations of purely reactive policies without temporal reasoning.  

LSTM baselines provide further perspective. The end-to-end LSTM has very low success (0.05), moderate collisions (0.15), long paths (mean 105.8), and slightly higher exposure (3.41), suggesting direct action-output models need more training to make strategic progress. The LSTM HL + MPC variant improves success (0.30) and time-to-goal when successful (16.6) with longer paths (56.2), but shows high collision (0.70) and higher exposure (3.49). This indicates hierarchical decomposition can aid efficiency, but without stronger LSTM training, temporal reasoning and visibility-aware subgoal selection remain weaker than HAVEN.

\textcolor{black}{Comparing HAVEN to recent RL-based covert path planning approaches such as EnCOMP \cite{hossain2024encomp} provides additional insight. EnCOMP has a significantly lower success rate (0.42) with collision rates comparable to HAVEN, but exhibits a higher average enemy exposure time (3.67). While its overall task completion is competitive, the increased exposure suggests weaker visibility-aware planning and less consistent covert behavior compared to HAVEN.}

Overall, these results indicate that:  
(1) memory is essential for risk-aware planning under partial observability;  
(2) visibility-aware features and candidate masking provide an effective inductive bias; and  
(3) hierarchical decomposition reduces adversarial exposure while avoiding the instability of end-to-end reactive control.  

\section{Discussion}
The DTQN’s short-horizon attention mitigates observation aliasing, enabling consistent exploitation of occlusion and reducing adversarial exposure without compromising goal progress. Restricting learning to high-level subgoal selection avoids the large state-space complexity of end-to-end control in cluttered, adversarial maps. Visibility-aware features and candidate pruning further focus learning on tactically relevant trade-offs (e.g., cover vs.\ exposure) rather than low-level geometric details.  

This decomposition also eases transfer: the 2D–3D extension reuses the same feature schema via planar point-cloud projection, requiring no changes to the high-level architecture. In 3D Unity--ROS experiments, behavior remains robust despite sensing noise, aided by normalization and persistence filtering. A noted limitation is occasional unsmooth motion due to latency between perception and control, suggesting the need for latency-aware feedback mechanisms.  

\section{Future Work}
Future directions include:
\begin{itemize}
    \item \textbf{Adversarial dynamics:} explicitly model enemy intent and extend to multi-agent cooperative settings.
    \item \textbf{Path corridor reasoning:} augment candidate features with predicted path corridors (minimum distance to FoVs, expected exposures, clearance times).
    \item \textbf{Real-world deployment:} real-robot experiments with closed-loop control and robust perception under sensor noise and unpredictable adversarial policies.
    \textcolor{black}{\item \textbf{3D-native perception:} replace the planar projection pipeline with learned 3D-to-feature encoders to handle multi-level environments and non-planar occlusion.}
\end{itemize}

\section{Conclusion}
We introduced a hierarchical, memory-augmented framework for adversarial navigation, coupling a DTQN-based subgoal selector with a low-level potential field controller. The design leverages visibility-aware features and candidate masking to achieve higher success, safety, and stealth relative to classical planners and baselines, while remaining competitive in efficiency. 
\textcolor{black}{Key formulations: Double DQN for stable Q-value estimation, visibility-aware candidate masking for safety-biased exploration, and subgoal hysteresis for trajectory smoothness all benefit overall performance, as validated by training stability analysis and ablation studies.}

Importantly, the approach transfers directly from 2D to 3D via point-cloud projection, demonstrating architectural flexibility and practical scalability. Comparisons against baselines and ablations highlight the benefits of temporal memory and hierarchical decomposition for navigation under partial observability. These findings suggest a promising path toward safe and reliable autonomous navigation in adversarial, visibility-constrained environments.

\section*{ACKNOWLEDGMENTS}
This material is based upon work supported in part by the DEVCOM Army Research Laboratory under cooperative agreement : W911NF2520170.

{\small
\bibliographystyle{IEEEtran}
\bibliography{references}
}

\end{document}